\documentclass[letterpaper, 10 pt, conference]{ieeeconf}  

\IEEEoverridecommandlockouts                              

\usepackage{amsmath} 
\usepackage{booktabs}
\usepackage{amssymb}  
\usepackage{graphicx}
\usepackage{xcolor}
\usepackage{changes}
\usepackage{multirow}
\usepackage{tabularx}
\usepackage{titlesec}

\title{\LARGE \bf
Decision-Aware Uncertainty Evaluation of Vision-Language Model-Based Early Action Anticipation for Human-Robot Interaction}
\newif\ifanonymous
\anonymousfalse 

\ifanonymous
\author{
Anonymous Authors%
\thanks{}%
\thanks{}%
\thanks{}%
\thanks{}%
}
\else
  \author{
    Zhaoda Du$^{1}$,
    Michael Bowman$^{2}$,
    Qiaojie Zheng$^{1}$,
    and Xiaoli Zhang$^{1}$%
    \thanks{$^{1}$Zhaoda Du, Qiaojie Zheng, and Xiaoli Zhang are with the Colorado School of Mines, Golden, CO 80401, USA
    (e-mail: zhaoda\_du@mines.edu; zheng@mines.edu; xlzhang@mines.edu).}%
    \thanks{$^{2}$Michael Bowman is a Postdoctoral Fellow in the Cancer Biology Department at the University of Pennsylvania, Philadelphia, PA 19104, USA
    (e-mail: Michael.Bowman@Pennmedicine.upenn.edu).}%
    \thanks{Corresponding author: Xiaoli Zhang (xlzhang@mines.edu).}%
    \thanks{}%
  }
\fi
\begin{document}

\maketitle
\thispagestyle{empty}
\pagestyle{empty}

\begin{abstract}
Robots in shared workspaces must interpret human actions from partial, ambiguous observations, where overconfident early predictions can lead to unsafe or disruptive interaction. This challenge is amplified in egocentric views, where viewpoint changes and occlusions increase perceptual noise and ambiguity. As a result, downstream human-robot interaction modules require not only an action hypothesis but also a trustworthy estimate of confidence under partial observation. Recent vision-language model-based approaches have been proposed for short-term action recognition due to their open-vocabulary and context-aware reasoning, but their uncertainty reliability in the temporal-prefix regime is largely uncharacterized. We present the first systematic evaluation of uncertainty in vision-language model-based short-term action recognition for human-robot interaction. We introduce a temporal-prefix evaluation protocol and metrics for calibration and selective prediction. We also characterize miscalibration patterns and failure modes under partial observations. Our study provides the missing reliability evidence needed to use vision-language model predictions in confidence-gated human-robot interaction modules.

\end{abstract}

\section{INTRODUCTION}

Robots that collaborate with people must interpret human actions from partial, evolving observations and do so early enough to be useful. In shared workspaces, early evidence is often ambiguous. Deciphering intent from initial motion is challenging and continues to be an open problem in human robot collaboration  \cite{hoffman2024intent}. Committing too early for an action can lead to disruptive or unsafe robot behavior. Likewise, robots should not be paralyzed by inaction while waiting for more confirming evidence.

Recently, vision-language models (VLMs) have emerged as an ingredient for action understanding in unstructured environments because they can leverage visual context and flexible language descriptions \cite{pani2025gazevlm}. However, for short-term (early/prefix) action prediction, accuracy alone is not sufficient to judge readiness for human-robot interaction (HRI) integration. Downstream HRI modules often depend on a decision-relevant confidence signal, an action hypothesis accompanied by a reliability estimate that reflects uncertainty under partial observation. Without evidence that VLM confidence is reliable in the prefix regime, and under additional noise such as egocentric viewpoints, it is difficult to use these predictions as inputs to confidence-gated HRI control systems.
\begin{figure}[!t]
\centering
\includegraphics[width=0.98\linewidth]{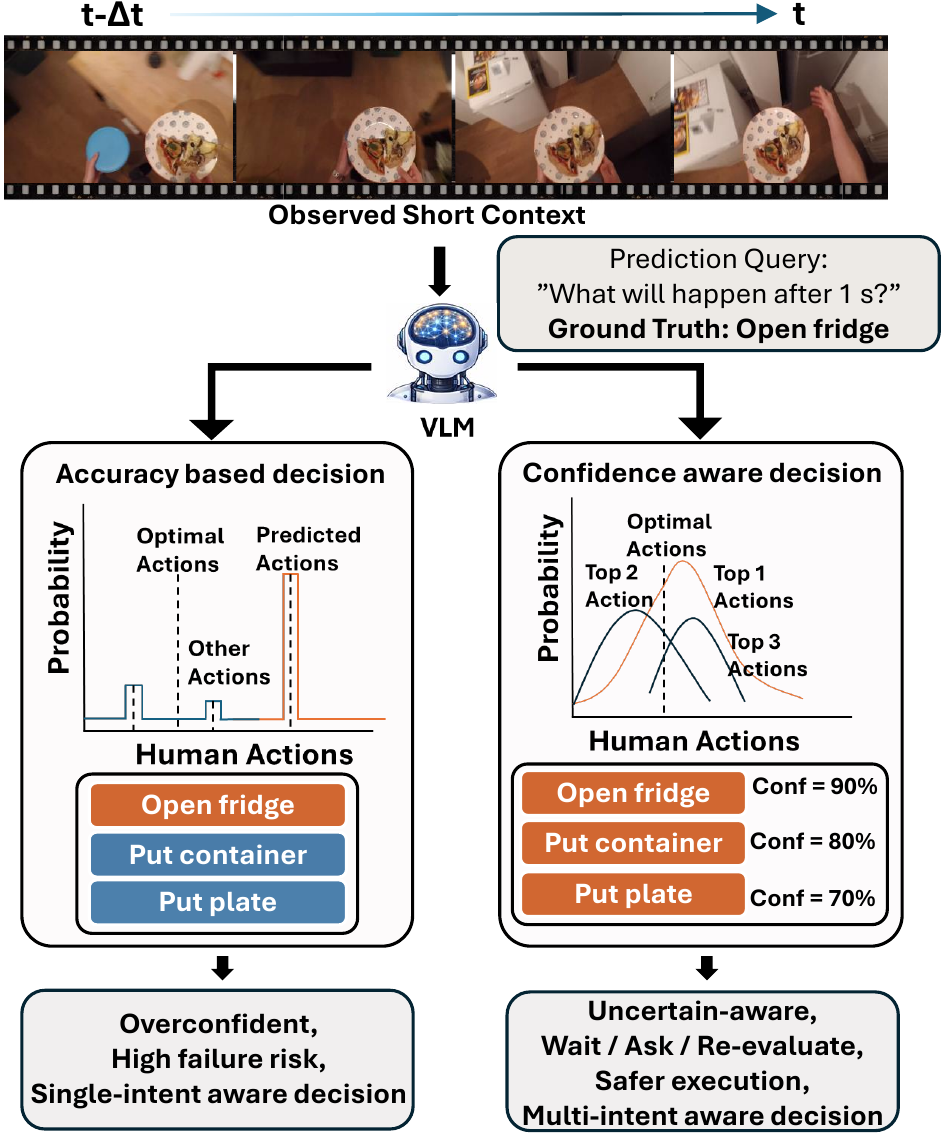}

\caption{Comparison between traditional accuracy-based Top-K selection and our confidence-aware selection framework for short-term action anticipation. Given observed short context, a VLM produces Top-K action hypotheses. Traditional methods select the highest-scoring action and commit early to a single hypothesis. In contrast, our approach models confidence and preserves multiple plausible intents, enabling delayed commitment, active clarification, and uncertainty-calibrated execution.}
\label{fig:tradeoff}
\end{figure}

We present the first systematic evaluation of uncertainty in VLM-based short-term action prediction for HRI. We introduce a temporal-prefix evaluation protocol and metrics for calibration and selective prediction. We also characterize miscalibration patterns and failure modes under partial observations. Our evaluation is grounded in established reliability methodology from uncertainty quantification and selective prediction \cite{geifman2017selective,zheng2025enhancing,he2025survey,traub2024selective,zhou2024novel}, and is motivated by the central role of uncertainty-aware decision making in HRI \cite{hoffman2024intent,lidard2024risk,hu2022active}. Our aim is to empower VLM predictions with an additional confidence score to eventually be used for downstream uncertainty-aware control modules.



The contributions of this work are threefold:

1) We reframe Top-$K$ short-term action anticipation as a reliability problem rather than purely a ranking problem, highlighting the need for uncertainty evaluation under partial observation in HRI settings.

2) We introduce a decision-aware evaluation framework for Top-$K$ outputs, including correctness, uncertainty reliability, selective decision utility, and confidence geometry analysis, providing practical tools to assess whether confidence signals are suitable for confidence-gated interaction.

3) Through systematic empirical analysis, we reveal that aggregation strategies fundamentally reshape the geometry of uncertainty, inducing trade-offs between calibration fidelity and decision-level separability. Our results demonstrate that improved ranking performance does not necessarily imply improved uncertainty reliability.




\begin{figure*}[t]
\centering
\includegraphics[width=\linewidth]{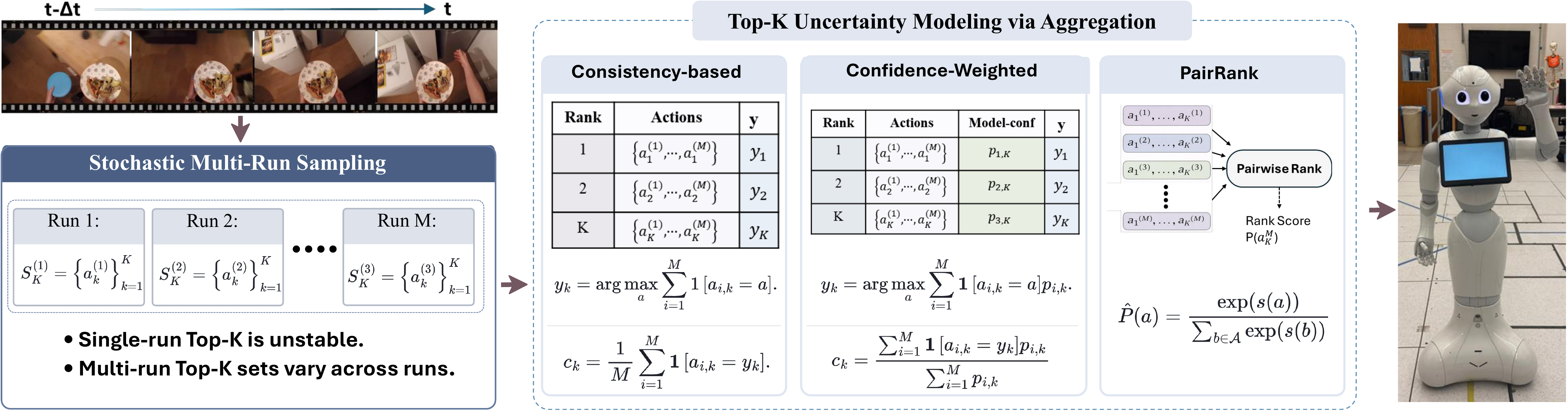}

\caption{From stochastic variability to structured uncertainty for HRI decision-making.
Multiple stochastic decoding runs produce unstable Top-K action hypotheses under partial observation. Aggregation operators transform these discrete prediction sets into a structured confidence distribution over candidate actions. Different aggregation strategies reshape the geometry of confidence in distinct ways, influencing calibration behavior and confidence-based decision gating in downstream human–robot interaction systems.}
\label{fig:aggregation_overview}
\end{figure*}

\section{Related Work}

\subsection{VLM-Based Early and Egocentric Action Understanding under Partial Observation}



Recent VLMs have enabled open-vocabulary early action prediction in egocentric settings, reporting competitive Top-1 and Top-K performance under partial observation \cite{pani2025gazevlm,kim2024palm,zhang2023video}. However, existing work primarily evaluates ranking accuracy.

In prefix-based anticipation, visual evidence is incomplete and multiple candidate intents may remain plausible. In HRI, the key issue is therefore not only which action ranks highest, but whether confidence meaningfully reflects ambiguity during early observation \cite{bowman2019intent,li2020inference}.

Although conventional classifiers provide Top-K probabilities via softmax outputs \cite{furnari2018leveraging}, VLM-based early action prediction differs structurally: the output space is language-driven, prefix ambiguity induces multi-intent uncertainty, and confidence signals are not guaranteed to align with decision-level reliability. The uncertainty behavior of Top-K predictions in this regime remains largely unexplored.
\subsection{Uncertainty-Aware Decision Making in HRI}

Uncertainty is fundamental in HRI,
as human intent and future motion are only indirectly observable.
Recent robotics surveys highlight intent inference and action prediction
as central capabilities for collaborative systems~\cite{hoffman2024intent,10582423,dragan2013legibility}.
Complementary works incorporate uncertainty into planning and interaction,
including risk-aware decision making and active uncertainty reduction~\cite{lidard2024risk, hu2022active,nikolaidis2017human,sadigh2016planning}.

However, these approaches typically assume that upstream perception modules
provide reliable uncertainty signals.
Our focus is narrower;
instead of proposing a new interactive planner, we evaluate the precursor needed to improve downstream uncertainty-aware decision modules. Specifically, we evaluate the reliability and mechanisms that produce confidence estimates in VLM-based short-term action recognition.

\subsection{Uncertainty Evaluation}

Calibration and selective prediction have been widely studied in supervised classification~\cite{minderer2021revisiting,he2025survey,traub2024selective}, and more recently in VLMs under classification or retrieval settings~\cite{xia2025survey,kuhn2023semantic,kadavath2022language}. Metrics such as Expected Calibration Error (ECE) and accuracy-rejection curve \cite{lin2024generating, guo2017calibration,el2010foundations,chow2003optimum} evaluate whether scalar confidence estimates align with empirical correctness and support confidence-based abstention.

However, these evaluations typically assume single-label correctness and do not explicitly address early multi-intent Top-$K$ decision regimes. In prefix-based action anticipation for HRI, the relevant evaluation target extends beyond Top-1 accuracy to \emph{set-level validity} and the structural distribution of confidence across ranked hypotheses. Moreover, collaborative control modules may reason over multiple candidate actions prior to execution~\cite{hoffman2024intent,lidard2024risk,hu2022active}, making confidence structure directly relevant to downstream planning.

Rather than introducing entirely new probabilistic measures, we reorganize uncertainty evaluation around decision-relevant properties for HRI, including candidate validity, set-level calibration, selective execution behavior, and confidence geometry. This perspective shifts the focus from purely probabilistic fidelity to interaction-aware reliability in early VLM-based action prediction.
\section{Method}
\subsection{Uncertainty Generation: Stochastic Multi-Run Sampling}

Uncertainty in predictive models is commonly characterized through a predictive distribution over possible outputs. However, in VLMs, direct access to such distributions is typically unavailable, as internal logits or probabilistic representations are not exposed. To approximate predictive uncertainty without modifying the model architecture, we leverage stochastic decoding as an implicit sampling mechanism.

As shown in Fig.~\ref{fig:aggregation_overview}, given an input egocentric video clip $x$, we perform $M$ independent stochastic forward passes under identical prompts and decoding configurations (e.g., temperature-based sampling). Each run produces a Top-$K$ action prediction set, $S_K$, for actions, $a^{(i)}_K$:
\begin{equation}
S_K^{(i)} = \{a^{(i)}_1, \dots, a^{(i)}_K\}, \quad i=1,\dots,M.
\end{equation}

The variability across these stochastic Top-$K$ outputs serves as a proxy for model uncertainty. Intuitively, if the model is confident about a particular action set, repeated stochastic decoding should yield highly consistent predictions. Conversely, significant variation across runs indicates increased uncertainty under the given input.
\subsection{Aggregation of Stochastic Top-$K$ Predictions}

As shown in Fig.~\ref{fig:aggregation_overview}, Given the stochastic Top-K outputs defined in Sec. III-A,
we aggregate them into a single ranked prediction
$\hat{S}_K = \{y_1, \dots, y_K\}$ with associated confidence scores.

We study three aggregation strategies \cite{xiong2023can}, summarized in Table~\ref{comparison_aggregation} and described below.
\begin{table}[!ht]
\centering

\caption{Comparison of aggregation strategies.}
\label{comparison_aggregation}
\begin{tabular}{lccc}
\toprule
Method & Rank-wise & Verbalized Conf. & Global \\
\midrule
Consistency & \checkmark & $\times$ & $\times$ \\
Confidence-weighted & \checkmark & \checkmark & $\times$ \\
Pairwise Ranking & $\times$ & $\times$ & \checkmark \\
\bottomrule
\end{tabular}
\end{table}

\subsubsection{Consistency-Based Aggregation}

We first construct the aggregated Top-$K$ prediction based on empirical agreement across stochastic runs.

Let $a_{i,k}$ denote the k-th ranked action in run i
(as defined in Eq. (1)).
For each position $k$, the aggregated action is determined by majority voting:
\begin{equation}
    y_k =
    \arg\max_{a}
    \sum_{i=1}^{M}
    \mathbf{1}[a_{i,k}=a].
\end{equation}

The corresponding confidence signal at position $k$ is defined as the agreement frequency
\begin{equation}
    c_k =
    \frac{1}{M}
    \sum_{i=1}^{M}
    \mathbf{1}[a_{i,k}=y_k].
\end{equation}


\subsubsection{Confidence-Weighted Aggregation}

Unlike the previous strategy, this approach incorporates model-reported verbalized confidence to account for additional uncertainty information.

For each run, the model reports a confidence $p_{i,k}\in[0,1]$ for action
$a_{i,k}$ at rank $k$. We compute a confidence-weighted vote at each position:
\begin{equation}
    y_k =
    \arg\max_{a}
    \sum_{i=1}^{M}
    \mathbf{1}[a_{i,k}=a]\;p_{i,k}.
\end{equation}
The associated confidence at rank $k$ is defined as the normalized support:
\begin{equation}
    c_k =
    \frac{
        \sum_{i=1}^{M}\mathbf{1}[a_{i,k}=y_k]\;p_{i,k}
    }{
        \sum_{i=1}^{M} p_{i,k}
    }.
\end{equation}
\subsubsection{Pairwise Ranking Aggregation (PairRank)}

The previous aggregation strategies operate independently at each rank.
In contrast, this approach models global ranking structure
among candidate actions.

Let $\mathcal{A}$ denote the set of unique actions
appearing across stochastic Top-$K$ outputs.
For each run, if action $a$ is ranked above $b$,
we record a pairwise preference event $a \succ b$.
Aggregating all such events yields a set of pairwise comparisons.

We estimate latent utility scores $s(a)$
by fitting a Bradley-Terry model via maximum likelihood estimation\cite{bradley1952rank}.
Normalized probability distribution over actions is then defined as:
\begin{equation}
    \hat{P}(a)
    =
    \frac{\exp(s(a))}
    {\sum_{b \in \mathcal{A}} \exp(s(b))}.
\end{equation}

The aggregated Top-$K$ prediction is obtained
by sorting actions according to $s(a)$,
and $\hat{P}(a)$ serves as the associated confidence signal.


\subsection{Evaluation Protocol}
\begin{table*}[t]
\centering
\caption{Decision-Aware Evaluation Framework for Top-K Uncertainty in HRI}

\label{tab:evaluation_structure}
\small
\renewcommand{\arraystretch}{1.15}

\begin{tabularx}{\textwidth}{l l X p{3.2cm}}
\toprule
\textbf{Attribute} & \textbf{Role in HRI} & \textbf{Verified Property} & \textbf{Metrics} \\
\midrule

Correctness
& Candidate validity
& Ranking quality of predicted Top-K actions
& Top-1 \newline Recall@K \\

\cmidrule(lr){1-4}

Uncertainty Reliability
& Probabilistic fidelity
& Alignment between confidence and empirical correctness
& Top-1 ECE  \newline Set-ECE vs $K$ \\

\cmidrule(lr){1-4}

Selective Decision Utility
& Decision gating
& Effectiveness of confidence-based abstention
& Coverage-Threshold \newline Accuracy-Threshold \\

\cmidrule(lr){1-4}

Confidence Geometry
& Ambiguity modeling
& Structural distribution of confidence mass
& Rank-wise Conf Distribution. \newline Normalized Entropy \\

\bottomrule
\end{tabularx}
\end{table*}
In confidence-aware HRI systems, action predictions directly affect downstream control decisions (e.g., execution, deferral, or clarification). Evaluation should therefore assess not only predictive correctness, but also the reliability and decision utility of the associated confidence signal.

We structure evaluation along four decision-relevant dimensions: correctness (candidate validity), reliability (calibration fidelity), selective utility (decision gating), and confidence geometry (ambiguity modeling), as summarized in Table~\ref{tab:evaluation_structure}.

\subsubsection{Correctness}

Correctness ensures that the predicted hypothesis set contains the true action, providing valid candidates for downstream planning. We first evaluate predictive correctness independent of uncertainty interpretation. This establishes whether different aggregation strategies alter ranking quality. 

We report Top-1 Accuracy, measuring whether the highest-ranked prediction matches the ground-truth action, and Recall@K, evaluating whether the ground-truth action appears within the predicted Top-K set.

In the short-term (prefix) regime, multiple future actions may remain plausible under partial observation. Recall@K therefore reflects ambiguity coverage under early evidence. However, these metrics assess ranking quality only and do not evaluate whether associated confidence scores are meaningful.

\subsubsection{Uncertainty Reliability}

Reliable calibration enables principled threshold selection in safety-critical interaction. 
Beyond ranking performance, a usable uncertainty signal must align with empirical correctness. 
We therefore evaluate calibration at both the \emph{single-hypothesis} and \emph{set} levels.

\textbf{Top-1 Calibration.} 
Top-1 Expected Calibration Error (ECE) measures the alignment between the confidence of the highest-ranked action and its empirical correctness, reflecting classical single-class calibration.

\textbf{Set-Level Calibration.} 
In multi-intent anticipation, the event of interest is whether the ground-truth action appears within the predicted Top-K set. 

Let the evaluation set contain $N$ input segments $\{x_j\}_{j=1}^{N}$ with ground-truth action label $y_j^{\ast}$.
For each segment $x_j$, the aggregated Top-$K$ prediction set is $\hat{S}_K^{(j)}=\{y_{j1},\dots,y_{jK}\}$ with confidence scores $\{c_{j1},\dots,c_{jK}\}$.
We define the set-correctness indicator as $z_j=\mathbf{1}\!\left(y_j^{\ast}\in \hat{S}_K^{(j)}\right)$.
The set-level confidence is defined as the mean confidence over the Top-$K$ set, $\bar{c}_j=\frac{1}{K}\sum_{k=1}^{K} c_{jk}$.

We compute Set-ECE by partitioning samples into $B$ bins $\{S_b\}_{b=1}^{B}$ based on $\bar{c}_j$ and measuring the discrepancy between average confidence and empirical set correctness:
\begin{equation}
\mathrm{Set\mbox{-}ECE}
=
\sum_{b=1}^{B}
\frac{|S_b|}{N}
\left|
\frac{1}{|S_b|}\sum_{j\in S_b} z_j
-
\frac{1}{|S_b|}\sum_{j\in S_b} \bar{c}_j
\right|.
\end{equation}

Lower Set-ECE indicates better alignment between predicted confidence and empirical Top-K correctness. 
To assess robustness under increasing ambiguity, we additionally report Set-ECE as a function of $K$.

This comparison distinguishes conventional single-class calibration from set-level uncertainty calibration under multi-intent anticipation in HRI.


\subsubsection{Selective Decision Utility}

In confidence-gated HRI systems, low-confidence predictions may trigger execution deferral, clarification, or delayed commitment instead of immediate action. We therefore evaluate whether the constructed uncertainty signal supports effective selective decision-making at the interaction level.


Since robotic execution ultimately requires committing to a single action hypothesis, we simulate a confidence-gated execution policy by selecting the highest-ranked prediction, i.e., $y_j^{\text{exec}} = y_{j1}$ with execution confidence $c_j^{\text{exec}} = c_{j1}$.

Given a confidence threshold $\tau$, the system executes the predicted action only if $c_j^{\text{exec}} \ge \tau$.
Coverage measures the fraction of inputs for which the system commits to execution:
\begin{equation}
\text{Coverage}(\tau)
=
\frac{1}{N}
\sum_{j=1}^{N}
\mathbf{1}\!\left(c_j^{\text{exec}} \ge \tau\right),
\end{equation}
Selective accuracy evaluates correctness over the retained subset:
\begin{equation}
\mathrm{Acc}(\tau)
=
\frac{
\sum_{j=1}^{N}
\mathbf{1}\big(c_j^{\mathrm{exec}} \ge \tau\big)\,
\mathbf{1}\big(y_j^{\mathrm{exec}} = y_j^{\ast}\big)
}{
\sum_{j=1}^{N}
\mathbf{1}\big(c_j^{\mathrm{exec}} \ge \tau\big)
}.
\end{equation}

A useful uncertainty signal should induce a monotonic trade-off: as $\tau$ increases, coverage decreases while selective accuracy increases. Strong threshold separability indicates that confidence effectively distinguishes reliable predictions from uncertain ones, supporting principled decision gating in HRI.

\subsubsection{Diagnostic Analyses: Confidence Geometry}
Confidence dispersion reflects ambiguity among competing intent hypotheses, informing interaction strategies.
To interpret differences in calibration and selective behavior, we analyze the internal geometry of Top-K confidence distributions.

\textbf{Rank-wise Confidence Distribution.} We examine how confidence mass is allocated across ranked hypotheses. 
Meaningful uncertainty should exhibit structured decay across ranks while avoiding excessive concentration on the top hypothesis, as prefix-based HRI settings often involve multiple plausible intents. This reflects a trade-off between discriminative sharpness and ambiguity preservation.

 \textbf{Normalized Entropy.} Normalized entropy characterizes the dispersion of confidence within the Top-K set. 
Low entropy indicates concentrated belief, while high entropy reflects distributed ambiguity. In prefix-based HRI settings, both overly flat and excessively concentrated distributions may be undesirable, as they respectively hinder action selection or risk overconfident commitment under partial observation.

To characterize the dispersion of confidence within the Top-$K$ set, 
we normalize the confidence scores for each input segment $x_j$ as 
$p_{jk} = \frac{c_{jk}}{\sum_{k=1}^{K} c_{jk}}$, for $k=1,\dots,K$.

The normalized entropy for segment $x_j$ is defined as
\begin{equation}
\tilde{H}_j
=
-
\frac{
\sum_{k=1}^{K} p_{jk} \log p_{jk}
}{
\log K
}.
\end{equation}
\begin{table*}[t]
\centering
\caption{Main results on Dataset A and Dataset B ($K=10$, bins=10).
}
\label{tab:main_results}
\small
\begin{tabular}{llccccc}
\toprule
Dataset & Method 
& Top-1 $\uparrow$
& Recall@10 $\uparrow$
& Top-1 ECE $\downarrow$
& Set-ECE@10 $\downarrow$
& Entropy  \\
\midrule

\multirow{4}{*}{EGTEA Gaze+}
& Pairwise Ranking & 0.151 & \textbf{0.487} & 0.6977 & 0.387 & \textbf{0.184} \\
& Consistency & \textbf{0.162} & 0.401 & 0.6009 & \textbf{0.027} & 0.958 \\
& Confidence-weighted & \textbf{0.162} & 0.403 & 0.6175 & 0.101 & 0.964 \\
& Single run & 0.160 & 0.425 & \textbf{0.2404} & 0.322 & 0.749 \\

\midrule

\multirow{4}{*}{EPIC-KITCHENS-100}
& Pairwise Ranking & 0.026 & \textbf{0.124} & 0.6377 & 0.027 & \textbf{0.419} \\
& Consistency & 0.030 & 0.114 & 0.5984 & 0.232 & 0.958 \\
& Confidence-weighted & \textbf{0.031} & 0.111 & 0.6259 & 0.273 & 0.972 \\
& Single run & 0.026 & 0.122 & \textbf{0.2495} & \textbf{0.022} & 0.891 \\

\bottomrule
\end{tabular}
\end{table*}

\section{Experiments Setup}

We evaluate uncertainty aggregation behavior for Top-$K$ short-term egocentric action anticipation under a black-box VLM setting. 
This section describes the dataset, model interface, sampling protocol, aggregation configuration, and evaluation implementation details.
\subsection{Datasets}

We evaluate on two standard egocentric action anticipation benchmarks:
EGTEA Gaze+\cite{li2018eye} and EPIC-KITCHENS-100\cite{damen2022rescaling}. EGTEA Gaze+ contains approximately 10K annotated first-person action segments 
with verb-noun labels in kitchen environments. EPIC-KITCHENS-100 is a large-scale egocentric dataset with over 90K action segments collected across diverse subjects and environments. Following standard short-term anticipation protocols, we use visual observations preceding action onset as input and evaluate prediction of the immediate next action. We follow the official train/test splits provided by each dataset and report results on the test sets.

\subsection{Vision-Language Models}
We evaluate a state-of-the-art VLMs accessed via black-box API:Gemini 2.5 Flash-lite~\cite{comanici2025gemini}. This model accept visual observations and natural-language prompts and return ranked action predictions in text form. Since our protocol requires multi-run stochastic sampling and Top-K extraction over large-scale datasets, we adopt a lightweight model variant to enable scalable uncertainty analysis under realistic deployment constraints.

\subsection{Sampling Protocol}

To expose stochastic variability, we perform \textit{self-random sampling} \cite{shi2024thorough}. 
For each input segment, we query the VLM $M$ times using identical prompts and temperature-based decoding with temperature $T = 0.8$. 
This setting introduces controlled stochasticity while preserving semantically plausible action predictions \cite{renze2024effect}. 
Each query produces a ranked Top-$K$ list. 
Unless otherwise specified, we use $K = 10$ and $M = 5$ sampling runs per input. 

\subsection{Evaluation Implementation}

Top-1 ECE and Set-ECE are computed using $B=10$ equal-width confidence bins over $[0,1]$. 

\section{Experimental results and discussion}
\begin{figure*}[htbp]
\centering
\includegraphics[width=\linewidth]{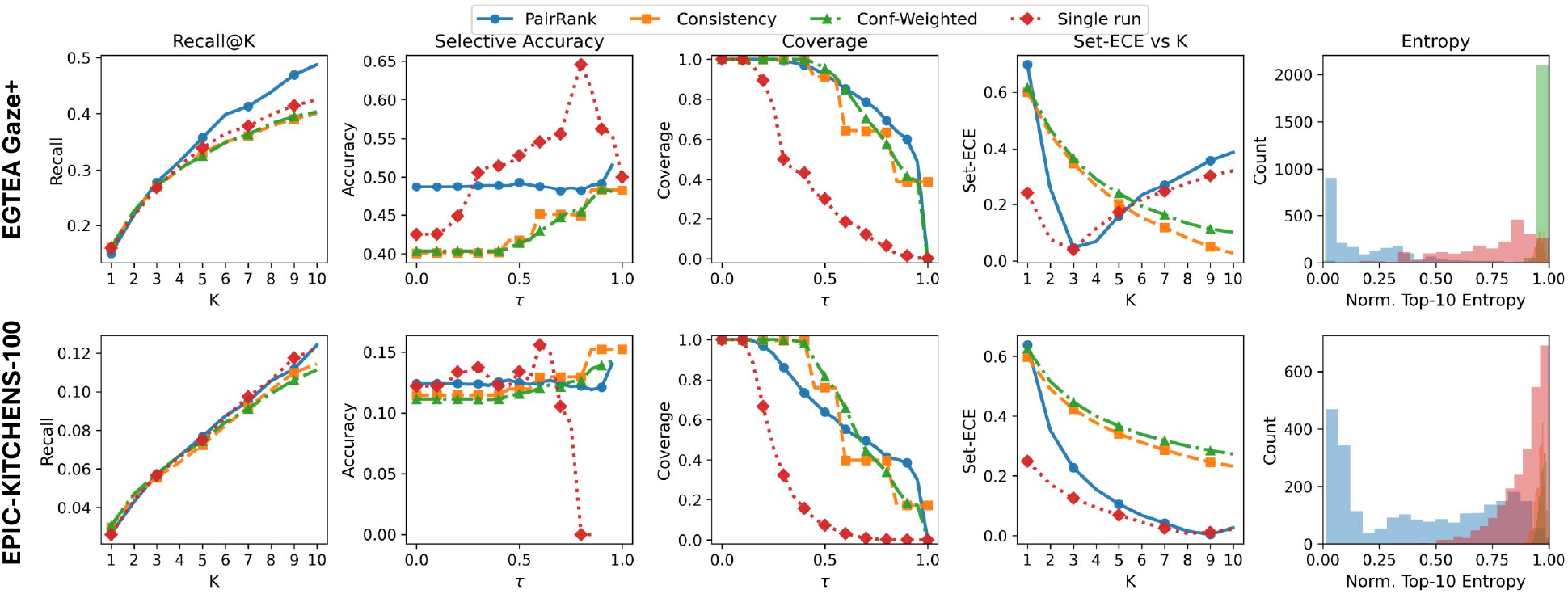}

\caption{
Comparison of aggregation strategies on two egocentric action benchmarks 
(top: EGTEA Gaze+; bottom: EPIC-KITCHENS-100). 
From left to right: Recall@K, selective accuracy and coverage under confidence thresholding 
, set-level calibration, 
and normalized Top-10 entropy. Ranking performance differences are modest. Calibration behavior and confidence geometry exhibit clearer structural differences across aggregation strategies, impacting selective decision characteristics.} 
\label{fig:all_figure}
\end{figure*}
Tables~\ref{tab:main_results} summarize quantitative metrics,
while Figures~\ref{fig:all_figure} and~\ref{fig:rankwise} illustrate ranking behavior, selective gating, calibration,
and confidence geometry across both datasets. For comparison, we include a single-run baseline, which directly uses the Top-$K$ prediction from one stochastic decoding pass without aggregation.

\subsection{Correctness}

Recall@K curves (Fig.~\ref{fig:all_figure}, first column) show that aggregation has
moderate impact on ranking performance.
In Table ~\ref{tab:main_results}, PairRank consistently achieves the highest Recall@10,
while Top-1 accuracy remains comparable across methods.
This suggests that aggregation primarily reshapes the ordering of candidates
without dramatically altering immediate prediction correctness.
\subsection{Uncertainty Reliability}

Calibration behavior differs substantially across strategies.
Top-1 ECE indicates that aggregation does not uniformly
improve probabilistic fidelity. In both datasets, the single-
run baseline achieves the lowest Top-1 ECE, while PairRank
exhibits the largest calibration error at $K=1$.

However, the Set-ECE vs.\ $K$ curves (Fig.~\ref{fig:all_figure}, fourth column)
reveal a more nuanced behavior. While PairRank shows
higher calibration error at small $K$, its Set-ECE decreases
more rapidly as $K$ increases. On EGTEA Gaze+, PairRank
achieves the lowest Set-ECE around $K=3$-$4$, whereas on
EPIC-KITCHENS-100 its calibration continues to improve
steadily with larger $K$.

This $K$-dependent behavior suggests that aggregation
strategies interact with the choice of decision horizon.
In HRI settings where multiple candidate actions are retained
prior to commitment, moderate $K$ values may yield
improved reliability under PairRank despite weaker Top-1
calibration. These findings highlight that calibration quality
should be interpreted jointly with the intended Top-$K$
decision regime rather than solely at $K=1$.
\subsection{Selective Decision Utility}
Selective accuracy–coverage curves (Fig.~\ref{fig:all_figure}, second and third columns)
demonstrate a clear separation effect.
PairRank maintains higher accuracy under increasing confidence thresholds
while reducing coverage more decisively.
This sharper threshold separability is especially visible
in the steep coverage decay of PairRank compared to averaging-based strategies.



This property is desirable in safety-critical HRI,
where decisive abstention may be preferable
to executing ambiguous predictions.
However, this benefit arises despite weaker Top-1 calibration,
highlighting a trade-off between calibration and selective decisiveness.
\begin{figure*}[htbp]
    \centering
    \includegraphics[width=0.7\linewidth]{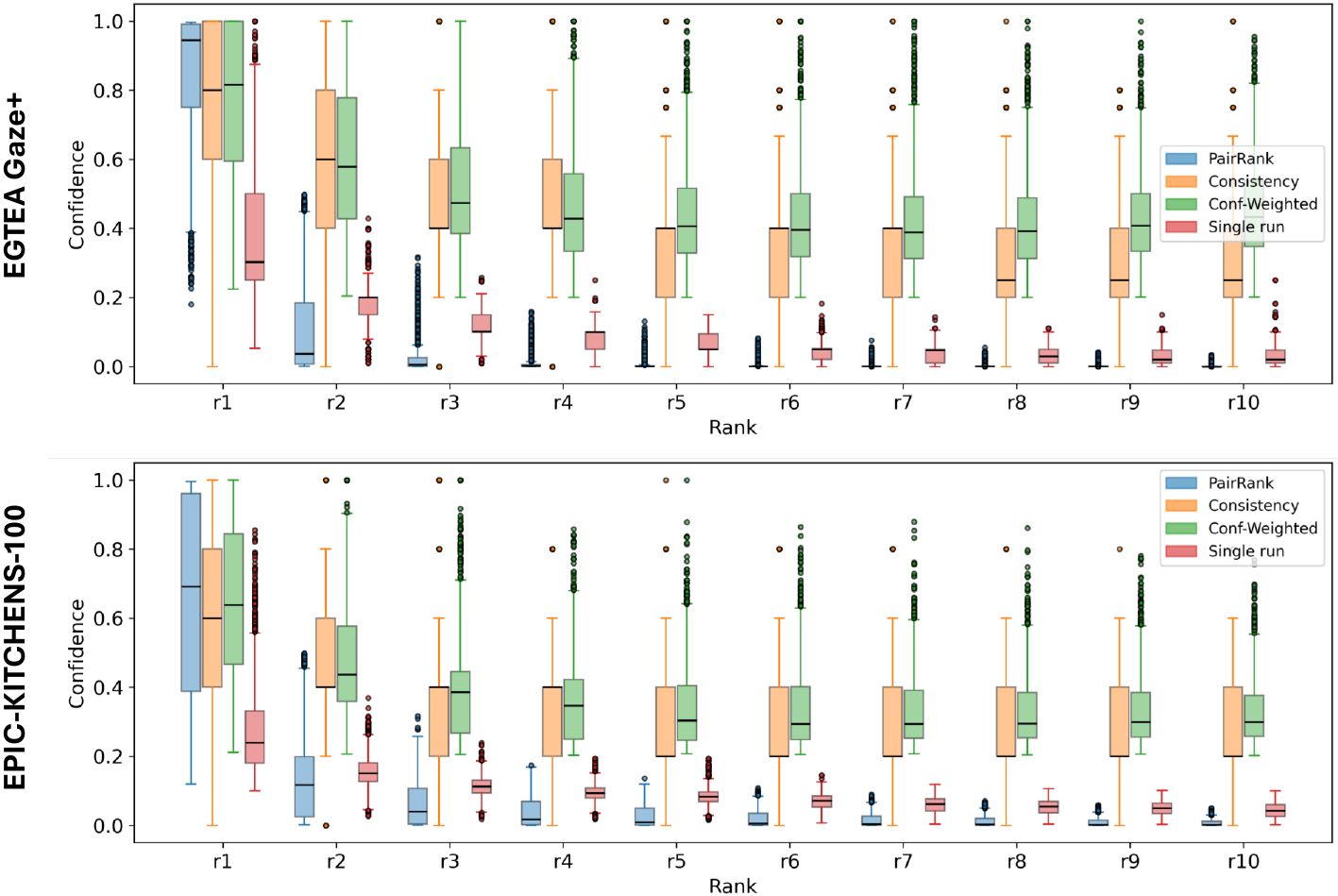}

    \caption{Rank-wise confidence distributions across aggregation strategies on EGTEA Gaze+ (top) and EPIC-KITCHENS-100 (bottom). Boxplots show the distribution of confidence assigned to each rank position within the Top-K set. PairRank sharply concentrates confidence on top ranks, whereas consistency and confidence-weighted methods yield smoother, higher-entropy distributions.}


  \label{fig:rankwise}
\end{figure*}
\subsection{Confidence Geometry}

Entropy distributions (Fig.~\ref{fig:all_figure}, fifth column) and
rank-wise confidence boxplots (Fig.~\ref{fig:rankwise})
provide structural insight into the observed behavior.

In Table~\ref{tab:main_results}, PairRank produces the lowest normalized Top-10 entropy
across both datasets, indicating strong concentration of probability mass.
The rank-wise boxplots further show that PairRank assigns
very high confidence to rank-1 predictions while sharply suppressing lower ranks,
creating a steep confidence hierarchy.
In contrast, consistency and weighted averaging
distribute confidence more evenly across ranks,
resulting in higher entropy and smoother confidence decay.

The structural differences in confidence highlight the crucial role uncertainty plays during interaction. For instance, low-entropy structures indicate a clear intent, whereas a high-entropy reflect ambiguous multi-intent states.
Such structural differences affect how a robot might
choose between execution, clarification, or information gathering.

\subsection{Interaction-Level Implications}

\begin{figure}[!t]
\centering
\includegraphics[width=\linewidth]{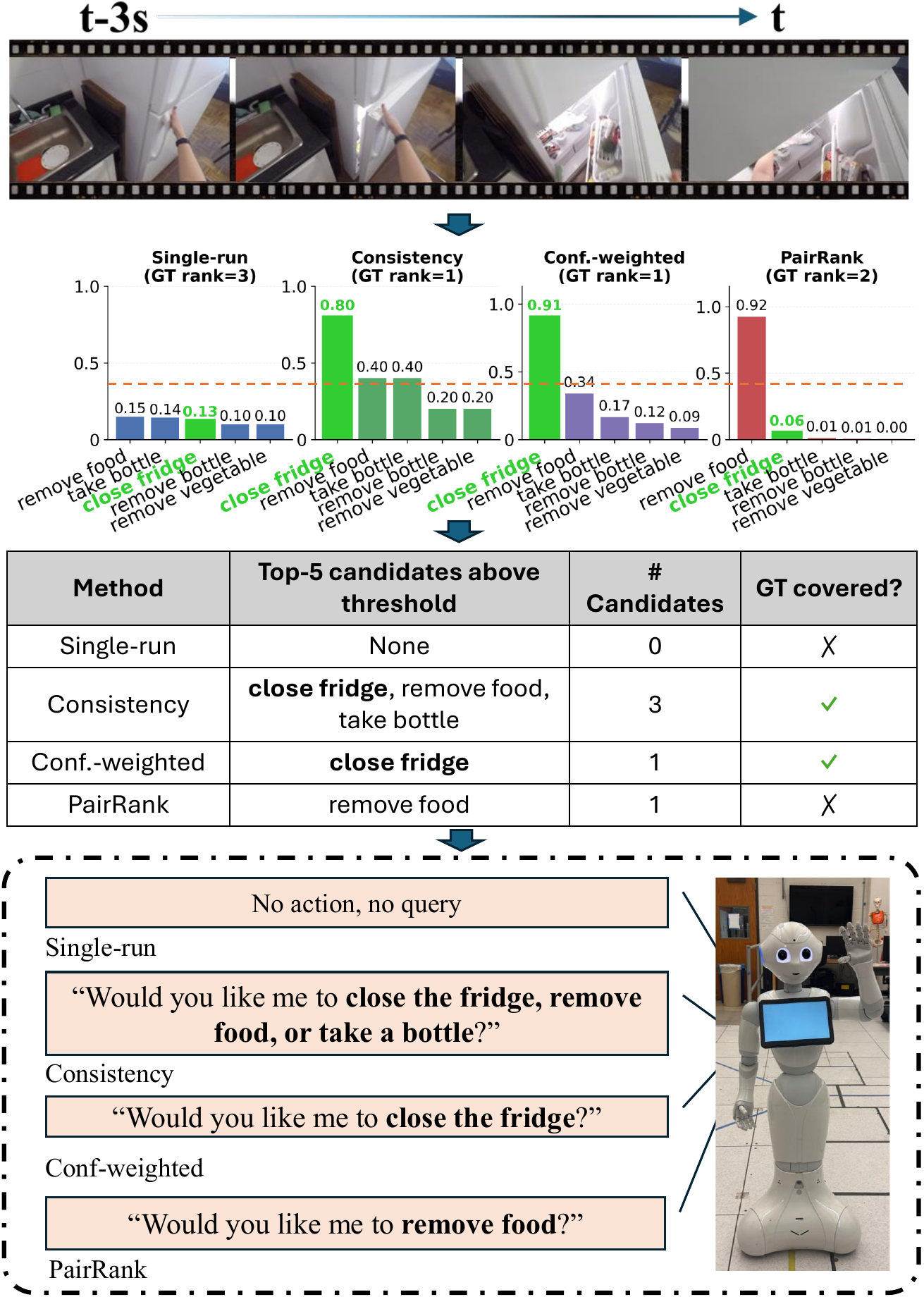}

\caption{
Illustrative example of Top-$K$ ($K=5$) prediction with confidence thresholding ($\tau = 0.39$) for a temporal prefix, used to examine interaction responses in HRI.
When only Top-$K$ rankings are considered, the aggregation strategies appear behaviorally similar.
However, incorporating confidence via threshold-based selection leads to markedly different clarification behaviors.
Sharp distributions (e.g., PairRank) may produce overconfident misprediction,
whereas smoother aggregations(e.g., Consistency) expand the clarification space and increase interaction burden.
Low-confidence single-run outputs may suppress interaction under thresholding.
This example highlights the importance of uncertainty-aware evaluation for interaction-level decision-making.
}

\label{fig:HRI}
\end{figure}

Figure~\ref{fig:HRI} illustrates that aggregation strategies cannot be evaluated solely by ranking performance.
The interaction-level behavior of the system depends critically on how Top-$K$ predictions are integrated with confidence-based decision rules, including the choice of $K$ and the confidence threshold $\tau$.
Different aggregation methods reshape the confidence geometry in distinct ways, which in turn determines clarification scope, interaction burden, and risk of overconfident misprediction.

Importantly, no aggregation strategy is universally optimal.
Sharp confidence distributions may enable efficient targeted confirmation but carry the risk of severe misjudgment when incorrect.
Smoother aggregations increase robustness by enlarging the candidate space, yet may introduce excessive clarification overhead in time-sensitive interactions.
Low-confidence outputs may suppress interaction entirely under threshold-based gating.

Therefore, the selection of aggregation strategy, Top-$K$ size, and confidence threshold should be guided by domain-specific requirements and expert-defined interaction constraints.
The appropriate trade-off between accuracy, reliability, and interaction complexity depends on the operational context of the human–robot system.
This highlights the necessity of uncertainty-aware evaluation beyond conventional Top-$K$ metrics. 
\subsection{Summary of Findings}

The experimental results demonstrate that aggregation does not merely affect ranking performance, but fundamentally reshapes uncertainty geometry. 
Although Top-$K$ accuracy may suggest comparable predictive capability, confidence distributions determine clarification scope, interaction risk, and system responsiveness. 
Sharp confidence can induce overconfident misprediction, while smoother distributions trade efficiency for robustness. 
Under threshold-based control, these structural differences translate directly into distinct human–robot interaction behaviors. 
Therefore, aggregation should be evaluated not only in terms of predictive correctness, but also with respect to reliability–efficiency trade-offs and interaction complexity. 
Uncertainty modeling thus emerges as a critical component of interaction-aware system design.


\section{Limitations and Future Work}
This study is centered on uncertainty evaluation rather than architectural performance optimization. Our objective is to highlight the importance of decision-aware uncertainty analysis in temporal-prefix action anticipation and to provide a structured evaluation framework for confidence behavior in HRI settings. As such, we do not aim to benchmark or compare multiple VLM architectures.

Experiments are conducted using a single representative black-box VLM because the contribution of this work lies in emphasizing the reliability of action predictions and the evaluation of uncertainty metrics, rather than in maximizing raw predictive performance. While testing additional large-scale VLMs could further validate cross-model consistency, such benchmarking falls outside the primary focus of this study. In practical deployment scenarios, the proposed framework can be directly applied to stronger or domain-adapted VLMs to achieve improved predictive performance while retaining the decision-aware uncertainty evaluation framework.

Hyperparameters including Top-K size, stochastic temperature, and sampling count are fixed for controlled comparison. Exploring adaptive or context-aware parameter selection remains an important direction for future work.

Furthermore, the interaction analysis is based on simulated confidence-gated decision policies rather than real robotic deployment. Although the evaluation protocol captures principled trade-offs between coverage and selective accuracy, real-world HRI systems involve additional operational constraints and human feedback dynamics. Future work will integrate the proposed framework into closed-loop robotic systems to study real-time decision effects.
\section{Conclusion}

We presented a systematic evaluation of uncertainty behavior in temporal-prefix Top-K VLM-based action prediction for HRI. 
Our analysis shows that aggregation reshapes confidence geometry, inducing trade-offs between calibration fidelity and decision-level separability. 
These findings emphasize the need for decision-aware uncertainty evaluation in early multi-intent anticipation and provide empirical guidance for confidence-gated HRI integration.




\addtolength{\textheight}{-12cm}   






\bibliographystyle{IEEEtran}
\bibliography{IEEEabrv,refs}

\end{document}